\pdfoutput=1

\documentclass[11pt]{article}

\usepackage{EMNLP2022}

\usepackage{times}
\usepackage{latexsym}
\usepackage{titlesec}

\usepackage[T1]{fontenc}

\usepackage[utf8]{inputenc}

\usepackage{microtype}

\usepackage{inconsolata}

\usepackage{booktabs}
\usepackage{xcolor}         %
\usepackage{graphicx}
\usepackage{xspace}
\usepackage{todonotes}
\usepackage{multirow}
\usepackage{wrapfig}
\usepackage{caption}
\usepackage{subcaption}
\usepackage{tabularx}
\usepackage{amsmath}
\usepackage{bbm}
\usepackage{bm}
\usepackage{mathtools}
\usepackage{lipsum}
\usepackage[ruled,vlined]{algorithm2e}

\usepackage{pifont}%

\newcommand{\ie}{i.e.,\xspace}

\newcommand{\eat}[1]{}

\newcommand{\method}{Modular Transformers\xspace}

\usepackage{cleveref}

\title{Modular Transformers: Compressing Transformers into \\  Modularized Layers for Flexible Efficient Inference}

\author{Wangchunshu Zhou\thanks{~~Work done while interning at the Allen Institute for AI}\ \ \textsuperscript{1} \hspace{.3cm} 
\textbf{\hspace{.3cm} Ronan Le Bras} \textsuperscript{2} \hspace{.3cm}
\textbf{\hspace{.3cm} Yejin Choi} \textsuperscript{2~3} \\
\textsuperscript{1}ETH Zurich \\
\textsuperscript{2}Allen Institute for AI \hspace{.3cm}  \\
\textsuperscript{3}Paul G. Allen School of Computer Science \& Engineering, University of Washington \\
\texttt{wangchunshu.zhou@inf.ethz.ch}}

\begin{document}
\maketitle

\begin{abstract}
Pre-trained Transformer models like T5 and BART have advanced the state of the art on a wide range of text generation tasks. Compressing these models into smaller ones has become critically important for practical use. Common neural network compression techniques such as knowledge distillation or quantization are limited to \textit{static} compression where the compression ratio is fixed. In this paper, we introduce \method, a modularized encoder-decoder framework for flexible sequence-to-sequence model compression.
\method trains modularized layers that have the same function of two or more consecutive layers in the original model via module replacing and knowledge distillation. After training, the modularized layers can be flexibly assembled into sequence-to-sequence models that  meet different performance-efficiency trade-offs. Experimental results show that after a single training phase, by simply varying the assemble strategy, \method can achieve flexible compression ratios from 1.1$\times$ to 6$\times$ with little to moderate relative performance drop. 
\end{abstract}
\section{Introduction}
The ever increasing size of pre-trained sequence-to-sequence (seq2seq) models~\citep{lewis2019bart,raffel2019exploring,zhang2020pegasus,liu2020multilingual,xue2020mt5,zhou-etal-2021-improving-sequence}
has supported advances in 
the state of the art in a wide range of natural language processing (NLP) tasks. 
For example, BART~\citep{lewis2019bart} has 400 million parameters while T5~\citep{raffel2019exploring} pushes this number to 11 billion. 
This makes large pre-trained seq2seq models hard to deploy and prone to negative environmental impacts~\citep{strubell2019energy,schwartz2020green,xu2021survey}, motivating researchers to investigate methods to compress large pre-trained models into smaller, faster ones that retain strong performance. 
Previous work has shown that BERT~\citep{devlin2018bert}, a popular encoder-only pre-trained Transformer~\citep{vaswani2017attention}, can be effectively compressed and accelerated via different neural network compression techniques~\citep{sanh2019distilbert,sun2019patient,jiao2019tinybert,zhou2020bert,gordon2020compressing,shen2020q}. 

However, there is limited research on methods to 
compress pre-trained seq2seq models~\citep{shleifer2020pre}. 
On the other hand, pre-trained seq2seq models are generally more space and time-consuming compared to their encoder-only counterparts 
since they require storing the decoder as well 
and generally decode in an auto-regressive fashion. 
To satisfy ever-changing resource constraints varying in different applications and over time, existing seq2seq compression techniques must separately train and store many compact models with different compression ratios, which is computationally inefficient and may also violate space constraints. Moreover, as suggested by~\citep{kasai2020deep}, as opposed to encoder-only models, it is nontrivial to find proper sizes for seq2seq models with a given resource constraint as the depth for the encoder and the decoder must be jointly tuned. As such, searching for compact seq2seq models meeting different resource constraints can be very costly.
This motivates us to investigate the problem of \emph{flexible} seq2seq compression that can dynamically adjust compression ratio to meet varying resource constraints without training multiple compact models.

In this work, we present Modular Transformers, a framework to compress Transformers into modularized layers for flexible efficient inference. 
With the guidance from the original model, \method trains modularized layers that have the same function of different numbers of consecutive layers in the original model via multi-grained module replacing and knowledge distillation. 
Specifically, we first map each of the modularized Transformer layers to a sub-module (i.e., a number of consecutive layers) of the encoder or decoder of the original model. 
We then train the modularized layers by randomly assembling a model for each training step (replacing sub-modules of the original model with their corresponding modularized layers), while keeping the original model parameters frozen.
We propose a curriculum-replacing strategy to train the modularized layers in a fine-to-coarse fashion.
We also use attention and representation distillation to further encourage the modularized layers to behave similarly to the original sub-modules. 

After training, the modularized layers can be flexibly assembled into seq2seq Transformers that meet different performance-efficiency trade-offs. 
The compression ratio for encoders and decoders can also be seamlessly adjusted to find the optimal parameterization within a fixed computation budget without any additional training. In addition, we introduce two deterministic assembling strategies tailored for smaller sizes and lower latency. Both of them replace the original model following a decoder-to-encoder, top-to-bottom, and fine-to-coarse fashion. 

We empirically verify the effectiveness of \method by compressing T5~\citep{raffel2019exploring}, a popular pre-trained seq2seq model, on representative text generation tasks including text summarization, machine translation, and question generation. 
Empirical results show that our approach consistently outperforms prior seq2seq compression techniques across different datasets while also enabling users to flexibly adjust the efficiency-performance trade-off of the model to meet different requirements for deployment.


\begin{figure*}
    \centering
    \includegraphics[width=\textwidth]{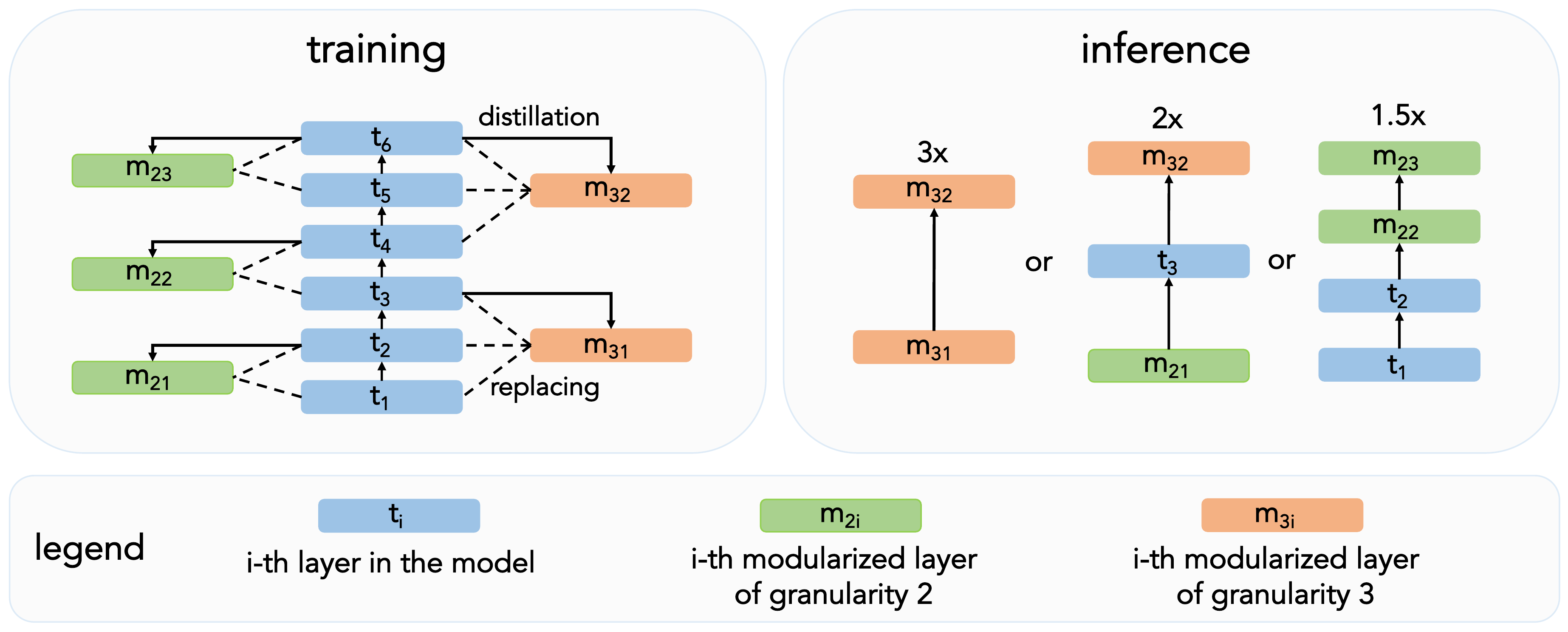}
    \caption{Illustration of the \method framework. A set of modularized layers with different granularity are trained by multi-grained module replacing and knowledge distillation. During inference, the modularized layers are assembled to meet different resource budgets.  $t_i$ denotes the i-th layer in the original model. $m_{ij}$ denotes j-th modularized layer with a granularity of i. 
    }
    \label{fig:my_label}
\end{figure*}

\section{Related Work}
\paragraph{Pre-trained Model Compression}
Prior work has shown that BERT~\citep{devlin2018bert}, a popular encoder-only pre-trained Transformer~\citep{vaswani2017attention}, can be effectively compressed and accelerated. As summarized in \citet{xu2021survey} and \citet{xu-etal-2021-beyond}, popular BERT compression techniques include knowledge distillation~\citep{hinton2015distilling,sanh2019distilbert,sun2019patient,jiao2019tinybert,zhou-etal-2022-bert} which trains a compact student network to mimic the behavior of the original teacher model, pruning~\citep{lecun1989optimal,michel2019sixteen,gordon2020compressing,sanh2020movement,wang2022efficientvlm} which prunes redundant neurons or structures in the original model, module replacing~\citep{xu2020bert} which trains compact successor sub-modules to replace that in the original model, and quantization~\citep{shen2020q,zafrir2019q8bert} that compresses a neural network by reducing the number of bits used to represent its parameters. In addition, a number of work also investigated efficient inference with BERT-like models with dynamic inference methods including early exit~\citep{teerapittayanon2016branchynet,xin2020deebert,liu2020fastbert,schwartz2020right,zhou2020bert} or adaptive computation time~\citep{graves2016adaptive,eyzaguirre2021dact}. These approaches only reduce inference latency while keep the model size unchanged. \method focuses on the first category which reduces the size of the model while also accelerating inference.

\paragraph{Sequence-to-sequence Compression} Compared to conventional neural network compression, seq2seq compression is relatively less explored. The de-facto method for compressing seq2seq model is sequence-level knowledge distillation~\citep{kim2016sequence}, which uses the original teacher model to generate pseudo-labels with beam search on the train set and train the compact student model with the input-pseuodo label pairs. A number of recent studies~\citep{junczys2019microsoft,kasai2020deep,zhang2021attention} have verified the effectiveness of the pseudo labeling method for distilling pre-trained Transformer-based seq2seq models. However, \citep{shleifer2020pre} shows that a simple shrink and fine-tune method performs competitively with pseudo labeling while requiring fewer computations. Another work related to \method is LayerDrop~\citep{fan2019reducing}, which randomly dropout Transformer layers when pre-training a seq2seq model and then prunes certain layers during inference for improved efficiency. LayerDrop differs from our method for two main reasons. First, LayerDrop must be applied \textit{during} pre-training while \method can be applied to any pre-trained models. This makes the scope of application of our method much broader because most existing models are not pre-trained with LayerDrop. Second, during inference, LayerDrop prunes layers while \method replaces sub-networks with compact modules with similar functionality. This hopefully reduces the performance drop, especially when the target compression ratio is relatively large. 

Moreover, there is also some prior work exploring pruning~\citep{michel2019sixteen,li2021differentiable} and quantization~\citep{li2022dq} techniques for seq2seq model compression. Recently, ~\citet{zhou2023efficient} proposed dynamic in-context learning for efficient text generation with prompting. These lines of work are orthogonal to knowledge distillation, pseudo labeling, and \method and the approaches can be combined in a straightforward manner. 



\section{Methodology}

In this section, we describe the proposed \method framework in detail. We first recap module replacing in \S\ref{para:bot}. We then describe the architecture design and training method for \method in \S\ref{para:modseq} and \S\ref{para:kd}, respectively. Then, we present the idea of dynamic assembling in \S\ref{para:assemble}. Finally, we discuss the relationship between \method and vanilla module replacing in \S\ref{para:relation}.

\subsection{Preliminary: Module Replacing}
\label{para:bot}

The goal of module replacing is essentially similar to knowledge distillation: training a compact model that behaves like the original large model. Compared to knowledge distillation which trains a student model to mimic a teacher model by minimizing the discrepancy between the predictions or hidden representations of the student and the teacher, the idea of module replacing is more direct: a compact model should behave the same way as the original model if all of its sub-networks are interchangeable with those in the original model.


Specifically, for a desired compression ratio $r$, we first specify a compact ``successor'' layer for each $r$ consecutive layer, which we refer to as sub-networks, in the original model. 
Consider a model $T$ with $n \times r$ layers, we can define a compact model $S$ which has $n$ layers. Let $T=\{t_1, \ldots, t_n\}$ denote the original model, $t_i$ and $s_i$ denote the sub-networks and layers in the original model and the compact model, respectively. 
The output vectors of the $i$-th sub-network or layer is denoted as $\mathbf{y}_i$.
During compression, in each step, we sample a random variable $r_{i+1}$ that determines whether the $(i+1)$-th sub-network in the original model is replaced by the corresponding compact layer in this step, from an independent Bernoulli distribution with a probability $p$ of success:
\begin{equation}
    r_{i+1} \sim \operatorname{Bernoulli}(p)
\end{equation}
As such, the output of the $(i+1)$-th model is calculated as:
\begin{equation}
    \mathbf{y}_{i+1} = r_{i+1} * s_i(\mathbf{y}_i) + (1 - r_{i+1}) * t_i(\mathbf{y}_i)
\end{equation}
where $*$ denotes the element-wise multiplication and $r_{i+1}\in \{0, 1\}$. In this way, the compact layers are trained to replace the corresponding sub-networks in the original model. After convergence, the compact layers are expected to be interchangeable (thus having the same functionality) with the original sub-networks. Moreover, the interaction between compact layers and original sub-networks and the random permutation of the hybrid model also adds extra regularization for the training of the compact model.



After training, we collect all compact layers and combine them to be the compressed model $S =\{s_1, \ldots, s_n\}$, 
which will be used for efficient inference.
Finally, we fine-tune the compact model to bridge the gap between module-replacing training and actual inference with the compact model.

\subsection{Modularized Seq2seq Models}
\label{para:modseq}

\method aims to train a set of modularized layers that can be flexibly assembled into compact models with different performance-efficiency trade-offs at test time. It can directly adapted to meet different resource constraints without re-training compact models of different sizes. To achieve this goal, we propose to define modularized layers with different capacities, \ie capturing the behavior of sub-networks of different sizes. 

Specifically, given an encoder (or decoder) $T=\{t_1, \ldots, t_n\}$ with $n$ layers and a target range of compression ratio from $1\times$ to $s\times$ (we assume n is divisible by s) , we define a suite of modularized layers $M=\{m_{ij}, i\in[n], j\in\{1, \ldots, n/i\}\}$, 
where $[n]$ denotes all positive integer divisors (or factors) of $n$ except 1 and $n$ itself, and $m_{ij}$ denotes a modularized layer that will be trained to match a sub-network that consists of $i$ consecutive layers starting from the $i\times (j-1)+1$-th layer in the original model. For example, for a Transformer model with 12 encoder/decoder layers and a target maximum compression ratio of 6, \method defines 6/4/3/2 modularized layers each corresponding to a sub-network representing 2/3/4/6 consecutive layers of the original model. Overall, \method consists of 15 modularized layers for the encoder as well as for the decoder, which is comparable to the original model size. After training, we can combine useful modularized layers into a compact model to reach a target compression ratio and a desired level of inference efficiency (see \cref{para:assemble}).

\subsection{Multi-grained Module Replacing}
\label{para:kd}

After defining the modularized layers in \method, we train them to have the same functionality as the original sub-networks via a combination of module replacing and knowledge distillation.

First, we extend the vanilla module replacing to multi-grained module replacing that can mix modularized layers with different granularity (\ie target sub-network size) during training. Since the target sub-networks of modularized layers of different granularity often overlap with each other, we can not simply sample a Bernoulli random variable to determine the structure of the hybrid model used for training in the current step. Instead, we propose a greedy replacing strategy for multi-grained module replacement. Specifically, we start from the n-th (last) layer of the original model and an empty hybrid model $H$. Assuming that we reach the k-th layer, we sample a random variable $r_{k} \sim \operatorname{Bernoulli}(p)$ where $p$ represents the probability of replacing that layer. If $r_{k} = 0$, we add $t_k$ into the hybrid model $H$ and move to the next layer. Otherwise, we perform a module replacement and sample a random variable $r_{ki} \sim \operatorname{Cat}(\bf{p})$ where $i \in [n]$ denotes the granularity of the modularized layer we want to add to the hybrid model. $\operatorname{Cat}(\bf{p})$ denotes a categorical distribution on $[n]$ (or Multinoulli distribution) and $\bf{p}$ is the probability vector of the $|[n]|$ possible outcomes. Then we determine the nearest suitable modularized layer of granularity $i$ as 
$m_{ij}, j = \lfloor k/i \rfloor$. 
If $i$ divides $k$, 
we can directly add $m_{ij}$ into $H$ and advance to the $k-i$-th layer. Otherwise, we first append the original layers $t_{i\times j+1:k}$  to $H$ before adding $m_{ij}$. In this way, for each training step, we sample a hybrid model and train its modularized layers with a task-specific objective (such as cross-entropy loss) to encourage the modularized layers to mimic the behavior of the replaced sub-networks in the original model.

In addition to module replacing, we also propose to leverage attention and hidden representation distillation~\citep{sun2019patient,jiao2019tinybert} to better align modularized layers and the corresponding sub-networks. Specifically, we train a modularized layer $m_{ij}$ in the hybrid model to match the attention distribution and the output hidden representation of $t_{i\times j}$ in the original model. During training, we simply combine task-specific loss with distillation loss with equal weights.

Moreover, we adopt the curriculum replacement strategy from~\citet{xu2020bert} and progressively increase the replacing probability $p$ to 1 during training. However, as opposed to vanilla module replacing where the model becomes static as $p=1$, the hybrid model is still randomly combined with modularized layers of different granularity. Also, motivated by the fact that modularized layers of larger granularity are more difficult to train, we propose to train the modularized layers of different granularity in a coarse-to-fine fashion. Specifically, for a 12-layer model (such that the modularized layers are of granularity in \{2,3,4,6\}), we start the training with $\textbf{p} = [1,0,0,0]$ so that only modularized layers of granularity $2$ will be sampled in the hybrid model. We then progressively (linearly) change $\textbf{p}$ to $[0.75,0.25,0,0]$, $[0.5,0.25,0.25,0]$, and finally to $[0.25,0.25,0.25,0.25]$ so that they are sampled uniformly. We empirically set each transition phase to a quarter of all training steps.

\subsection{Flexible Assembling}
\label{para:assemble}

We describe how we can use the trained modularized layers for flexible inference. The proposed flexible assembling method starts with the original model $T$ and gradually replaces it with modularized layers. In contrast to the random replacing strategy used for sampling the hybrid model during training, we need to take the target compression/speed-up ratio into account while also optimizing for the performance of the assembled model. To this end, we propose a deterministic decoder-first, top-down, fine-to-coarse replacing strategy for flexible assembling. The decoder-first and top-down strategy is inspired by the insight from prior work on seq2seq compression and module replacement. Specifically, \citet{shleifer2020pre} and \citet{kasai2020deep} revealed that within a fixed parameter budget, seq2seq models with a deep encoder and a shallow decoder generally perform the best, and \citet{xu2020bert} showed that top layers are more robust to module replacing, whereas replacing bottom layers often leads to a larger performance drop. The fine-to-coarse strategy is motivated by our conjecture that modularized layers with large granularity will more likely lead to a larger drop in performance.

In general, there exist two common compression strategies tailored for different kinds of resource budgets. The first setting focuses on the size of the compressed models and the second focuses on the inference speed. We devise two variants of our assembling strategy tailored for \textit{speed-first} and \textit{size-first} compression. Inspired by the previous observation of~\citet{shleifer2020pre}, we use a more uniform replacing strategy for size-first assembling while replacing decoder sub-networks more aggressively for speed-first assembling. 

We illustrate our approach on the T5-base model with $12$ encoder and $12$ decoder layers. In the size-first assembling, we first replace the decoder from top to bottom with modularized layers $m_{2j}$ corresponding to two consecutive layers. We then do the same for the encoder. After $T$ is entirely replaced by $m_{2j}$, we start replacing $m_{2j}$ with $m_{3j}$ from decoder to encoder, from top to bottom. We do this iteratively until the entire model consists only of $m_{6j}$ layers. As for speed-first assembling, we first replace decoder layers until they have all been replaced by $m_{4j}$ layers. We then replace the encoder with $m_{2j}$ and then replace the decoder with $m_{6j}$. Finally, we iteratively replace the encoder until the whole model consists only of $m_{6j}$ layers.

During this process, each module replacing operation reduces the number of layers in the model by $1$. As such, the compression ratio can be flexibly adjusted from $1$ to the maximum granularity of the modularized layers according to different resource constraints and performance-efficiency trade-offs.

\subsection{Relation with Module Replacing}
\label{para:relation}

Our approach differs from vanilla module replacing in four different ways: (1) module replacing is limited to compressing BERT-like encoder-only models in \citet{xu2020bert} while \method works with encoder-decoder models; (2) we propose to train modularized layers that can replace sub-networks of \textit{different sizes} in the original model and can be connected with others, whereas the successor layers are mapped to sub-networks of a fixed size; (3) we propose multi-grained module replacing and use representation distillation to better align modularized layers with their corresponding sub-networks while vanilla module replacing only supports a fix compression ratio and is trained with cross-entropy loss only; (4) we introduce the idea of dynamic assembling that enables \textit{flexible} adjustment of performance-efficiency trade-off, whereas the compression ratio in \citet{xu2020bert} is fixed since the successor layers are directly connected together. Essentially, we can view module replacing as a method used for training modularized layers so that they are interchangeable with sub-networks in the original model.

\begin{table*}[!tbp]
\centering
    \begin{minipage}{\textwidth}
	\centering
	\resizebox{\textwidth}{!}{
		\begin{tabular}{lcccccccccccc}
			\toprule
 & & &  \multicolumn{6}{c}{\textbf{Summarization}} & \multicolumn{3}{c}{\textbf{Question Gen}} & \textbf{Translation} \\
 & & & \multicolumn{3}{c}{\textbf{CNN/DM}} & \multicolumn{3}{c}{\textbf{XSum}} & \multicolumn{3}{c}{\textbf{Squad}} & \textbf{En-De} \\
\cmidrule(lr){4-6} \cmidrule(lr){7-9} \cmidrule(lr){10-12} \cmidrule(lr){13-13}  
\textbf{Method} & \textbf{\#Layers} &  \textbf{Ratio} & RG-1 & RG-2 & RG-L & RG-1 & RG-2 & RG-L & B-4 & M & RG-L & B-4 \\
\midrule
T5-base & 12-12 & 1.0/1.0$\times$ & 42.25 & 20.35 & 39.45 & 43.39 & 20.66 & 35.15 & 22.216 & 25.31 & 51.40 & 30.3  \\
\midrule
\multicolumn{13}{c}{\textit{Compressed Models Focusing on Smaller Sizes}} \\
\midrule
Pseudo Labeling & 6-6 & 1.5/2.0$\times$ & 41.08 & 19.24 &  38.15 & 42.34 & 19.91 & 34.09 & 21.36 & 24.52 & 50.55 & 29.1  \\
SFT & 6-6 & 1.5/2.0$\times$ & 41.16 & 19.48 & 38.24 & 42.21 & 19.46 & 33.83 & 21.24 & 24.45 & 50.42 & 28.5  \\
KD & 6-6 & 1.5/2.0$\times$ & 41.26 & 19.55 & 38.36 & 42.52 & 19.95 & 34.22 & 21.43 & 24.61 & 50.61 & 29.2 \\
Head Prune & 12-12 & 1.2/1.9$\times$ & 38.86 & 17.83 & 36.75 & 39.73 & 17.45 & 31.92 & 19.45 & 23.41 & 48.63 & 27.2 \\
LayerDrop & 6-6 & 1.5/2.0$\times$ & 37.45 & 16.38 & 35.62 & 38.62 & 16.61 & 30.78 & 18.86 & 22.82 & 47.25 & 25.1 \\
Quant + KD & 12-12 & 1.1/3.9$\times$ & 41.18 & 19.65 & 38.37 & 42.05 & 19.51 & 33.89 & 21.38 & 24.45 & 50.43 & 29.0 \\
\bf \method & 6-6 & 1.5/2.0$\times$ & \bf 41.71 & \bf 19.86 & \bf 38.83 & \bf 42.86 & \bf 20.27 & \bf 34.49 & \bf 21.53 & \bf 24.81 & \bf 50.81 & \bf 29.4  \\
\midrule
\multicolumn{13}{c}{\textit{Compressed Models Focusing on Lower Latency}} \\
\midrule
Pseudo Labeling & 12-3 & 1.9/1.6$\times$ & 41.25 & 19.38 & 38.27 & 42.49 & 20.01 & 34.26 & 21.40 & 24.45 & 50.47 & 29.3  \\
SFT & 12-3 & 1.9/1.6$\times$ & 41.45 & 19.63 & 38.55 & 42.41 & 19.74 & 33.96 & 21.38 & 24.52 & 50.51 & 28.8  \\
KD & 12-3 & 1.9/1.6$\times$ & 41.52 & 19.65 & 38.60 & 42.60 & 20.10 & 34.39 & 21.51 & 24.60 & 50.63 & 29.4 \\
LayerDrop & 12-3 & 1.9/1.6$\times$ & 38.31 & 17.37 & 36.44 & 39.44 & 17.23 & 31.46 & 19.28 & 23.31 & 47.93 & 25.9 \\
\bf \method & 12-3 & 1.9/1.6$\times$ & \bf 41.75 & \bf 19.88 & \bf 38.98 & \bf 42.99 & \bf 20.35 & \bf 34.59 & \bf 21.75 & \bf 24.95 & \bf 50.91 & \bf 29.6  \\
\bottomrule
	\end{tabular}
	}
	\caption{Static compression results with T5-base in both size-first (top) and speed-first (bottom) settings. For compression ratio, $a/b\times$ means the model is $a$ times faster and b times smaller than the original model. Head Prune and Quant + KD are only considered in the size-first setting, as they cannot be adjusted for size-first or speed-first and their speed-up ratio is relatively low.}
	\label{tab:resultbase}
	\end{minipage}
\end{table*}

\section{Experiments}

In this section, we empirically verify the effectiveness of \method by using it to compress T5~\citep{raffel2019exploring}, a popular pre-trained seq2seq Transformer, on a number of representative natural language generation tasks and datasets.

\subsection{Experimental Settings}

\subsubsection{Models and Datasets}

We conduct experiments with the T5-base and T5-large models, which have 220M and 774M parameters respectively, as the backbone models for base-size and large-size experiments.

As for tasks and datasets, we evaluate \method on text summarization, question generation, and machine translation. For summarization, we select the CNN/DailyMail~\citep{hermann2015teaching} and XSUM~\citep{narayan2018don} datasets. We use the SQUAD~\citep{rajpurkar2016squad} following the split in~\citet{du2018harvesting} for question generation and use the WMT-14~\citep{bojar2014findings} En-De split for machine translation.


\subsubsection{Baselines}

We compare \method with the following seq2seq compression methods: (1) \textbf{Pseudo Labeling (PL)}~\citep{kim2016sequence}, also called sequence-level knowledge distillation, which uses the original model to generate pseudo labeled data for training compact models; (2) \textbf{Shrink and Fine-tune (SFT)}~\citep{shleifer2020pre}, which simply shrinks the teacher model to student size and re-fine-tunes the shrunk model; (3) \textbf{Knowledge Distillation (KD)}~\citep{hinton2015distilling}, which combines logit distillation, hidden representation distillation, and attention distillation, to train a compact student model; (4) \textbf{Head Prune}~\citep{michel2019sixteen}, which prunes attention heads in the model with gradient-based head importance score; (5) \textbf{LayerDrop}~\citep{fan2019reducing}, which randomly drops Transformer layers during fine-tuning, and selectively prunes certain layers for efficient inference; and (6) \textbf{Quant + KD}~\citep{li2022dq}, which combines quantization-aware training and knowledge distillation to train a quantized seq2seq model.

\begin{table*}[!tbp]
\centering
    \begin{minipage}{\textwidth}
	\centering
	\resizebox{\textwidth}{!}{
		\begin{tabular}{lcccccccccccc}
			\toprule
 & & &  \multicolumn{6}{c}{\textbf{Summarization}} & \multicolumn{3}{c}{\textbf{Question Gen}} & \textbf{Translation} \\
 & & & \multicolumn{3}{c}{\textbf{CNN/DM}} & \multicolumn{3}{c}{\textbf{XSum}} & \multicolumn{3}{c}{\textbf{SQUAD}} & \textbf{En-De} \\
\cmidrule(lr){4-6} \cmidrule(lr){7-9} \cmidrule(lr){10-12} \cmidrule(lr){13-13}  
\textbf{Method} & \textbf{\#Layers} &  \textbf{Ratio} & RG-1 & RG-2 & RG-L & RG-1 & RG-2 & RG-L & B-4 & M & RG-L & B-4 \\
\midrule
T5-large & 24-24 & 1.0/1.0$\times$ & 42.61 & 20.72 & 39.81 & 43.83 & 21.05 & 35.44 & 22.75 & 25.45 & 51.62 & 31.2  \\
\midrule
\multicolumn{13}{c}{\textit{Compressed Models Focusing on Smaller Sizes}} \\
\midrule
Pseudo Labeling & 12-12 & 1.5/2.0$\times$ & 41.54 & 19.61 &  38.51 & 42.75 & 20.23 & 34.44 & 21.61 & 24.73 & 50.65 &  29.6  \\
SFT & 12-12 & 1.5/2.0$\times$ & 41.57 & 19.66 & 38.62 & 42.60 & 20.01 & 34.21 & 21.45 & 24.61 & 50.60 & 29.2  \\
KD & 12-12 & 1.5/2.0$\times$ & 41.62 & 19.73 & 38.70 & 42.83 & 20.21 & 34.55 & 21.59 & 24.75 & 50.71 & 29.9 \\
Quant + KD & 24-24 & 1.1/3.9$\times$ & 41.64 & 19.78 & 38.67 & 42.52 & 19.95 & 34.05 & 21.57 & 24.66 & 50.61 & 29.2 \\
\bf \method & 12-12 & 1.5/2.0$\times$ & \bf 41.92 & \bf 19.97 & \bf 39.01 & \bf 43.22 & \bf 20.51 & \bf 34.76 & \bf 21.79 & \bf 24.95 & \bf 50.86 & \bf 30.1  \\
\midrule
\multicolumn{13}{c}{\textit{Compressed Models Focusing on Lower Latency}} \\
\midrule
Pseudo Labeling & 24-6 & 1.9/1.6$\times$ & 41.76 & 19.79 & 38.68 & 42.92 & 20.28 & 34.51 & 21.72 & 24.94 & 50.81 &  30.1  \\
SFT & 24-6 & 1.9/1.6$\times$ & 41.85 & 19.95& 38.85 & 42.83 & 20.14 & 34.43 & 21.41 & 24.57 & 50.65 & 29.4  \\
KD & 24-6 & 1.9/1.6$\times$ & 41.92 & 20.01 & 38.94 & 42.97 & 20.35 & 34.53 & 21.45 & 24.64 & 50.70 & \bf 30.3 \\
\bf \method & 24-6 & 1.9/1.6$\times$ & \bf 42.12 & \bf 20.26 & \bf 39.25 & \bf 43.33 & \bf 20.68 & \bf 34.81 & \bf 21.79 & \bf 25.01 & \bf 50.95 & \bf 30.3  \\
\bottomrule
	\end{tabular}
	}
	\caption{Static compression results with T5-large. For compression ratio, $a/b\times$ means the model is $a$ times faster and $b$ times smaller than the original model. We only include methods that are competitive in the T5-base experiments.}
	\label{tab:resultlarge}
	\end{minipage}
\end{table*}

\subsubsection{Training Details}

We define modularized layers of granularity \{2,3,4,6\} for both T5-base and T5-large.\footnote{We do not include granularity 12 for T5-large because it leads to negative impact on overall performance, and a model with a compression ratio of 12 has very poor performance.} We train \method and all related models with a warm-up ratio of 0.05, a label smoothing~\citep{szegedy2016rethinking} rate of 0.1, and learning rate in \{3e-5, 5e-5, 7e-5\}.  For \method, we use a batch size of 128 and a max epoch of 24 for summarization and question generation datasets, and a batch size of 256 and a max epoch of 60 for machine translation datasets. We use the same batch size for all methods and the same number of epochs for all methods with KD. We train non-KD baselines with half the number of epochs which leads to similar performance and faster convergence. For method-specific hyperparameters, we adopt the values provided in the original contribution.

\subsubsection{Evaluation}

Following previous work, we report ROUGE~\citep{lin2004rouge} and BLEU~\citep{papineni2002bleu} for text summarization and machine translation, and report BLEU-4, METEOR~\citep{banerjee2005meteor}, and ROUGE-L for question generation. We compare \method with baseline methods in both \textbf{fixed} and \textbf{flexible} compression ratio scenarios. In the fixed budget scenario, we experiment with both the size-first and speed-first settings. The size-first setting focuses on the size of the compressed models and compresses both the encoder and the decoder by half, while the speed-first setting focuses on the inference speed and compresses the decoder by 3/4 but leaves the encoder uncompressed. The baselines are trained twice while \method can adjust to the two different trade-offs by simply varying the two aforementioned assembling strategies. In the flexible compression setting, we present both the trade-off between size-performance and speed-performance of \method and compare it with several common static compression settings trained with pseudo labeling.

\subsection{Experimental Results}

\subsubsection{Static Compression Results}

We present static compression results with T5-base and T5-large as the teacher model in Table \ref{tab:resultbase} and Table \ref{tab:resultlarge}, respectively. We can see that \method consistently outperforms all compared baselines in both size-first and speed-first settings without the need of re-training the model. We also find that while previous literature~\citep{kim2016sequence,shleifer2020pre} observes that logits-based knowledge distillation underperforms pseudo labeling and SFT baselines, adding attention distillation and hidden representation distillation makes KD performs slightly better than these baselines. Moreover, we find LayerDrop, the other baseline that does not require re-training for different compression ratios, performs poorly when the number of layers to be dropped is relatively large, which is consistent with previous observations~\citep{fan2019reducing}. Combining quantization and knowledge distillation performs well in terms of size-performance trade-off, which is consistent with a concurrent work by~\citet{li2022dq}. However, the speed-up of this approach is much smaller compared to other baselines, which 
we believe can be partially attributed to common GPUs not being optimized for quantized models.  
 
In addition, we find that models with 12-3 encoder/decoder layers consistently outperform their 6-6 counterparts while also leading to larger speed-ups. This is consistent with previous observations of~\citet{shleifer2020pre} and it confirms the effectiveness of the speed-first assembling strategy.

\begin{figure}
\centering
\begin{minipage}{.25\textwidth}
  \centering
  \includegraphics[width=\textwidth]{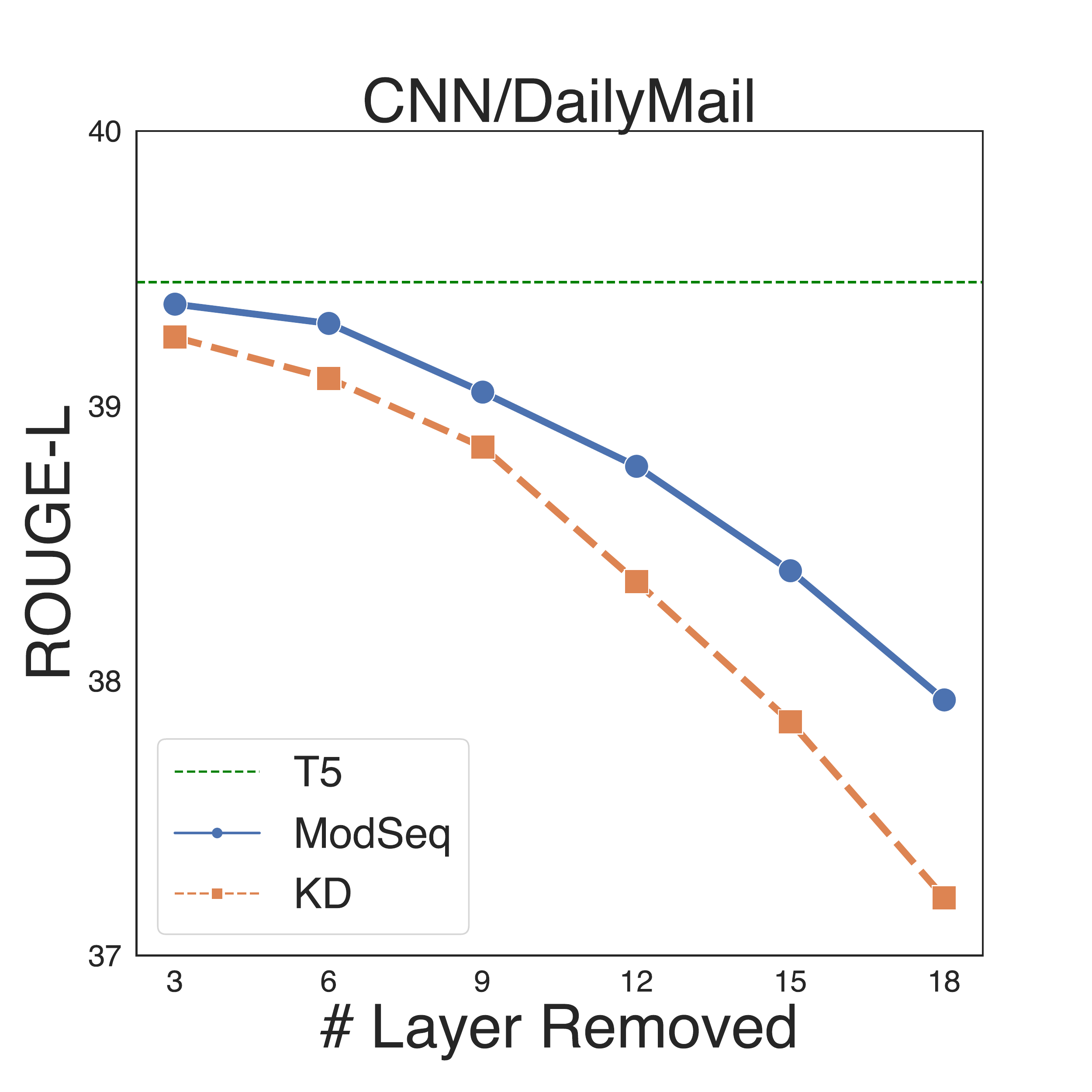}
  \label{fig:test1}
\end{minipage}%
\begin{minipage}{.25\textwidth}
  \centering
  \includegraphics[width=\textwidth]{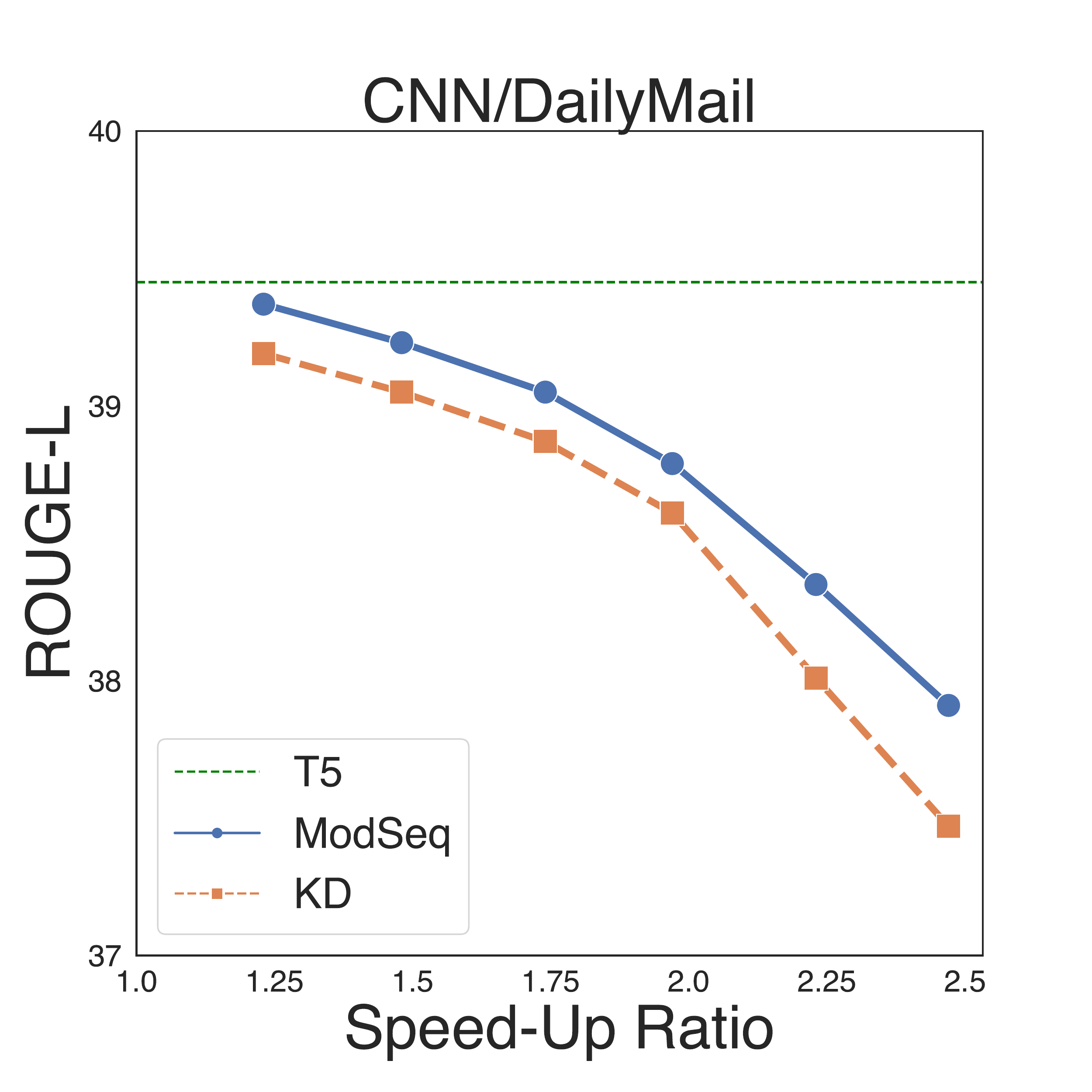}
  \label{fig:test2}
\end{minipage}
    \vspace{-0.6cm}
\caption{Flexible compression results of \method and KD with T5-base on the CNN/DailyMail dataset in both size-first (left) and speed-first (right) settings.}
\vspace{-0.4cm}
\label{fig:tradeoff}
\end{figure}

\subsubsection{Flexible Compression Results}

Similar to the static compression experiments, we also compare \method with the KD baseline, which is the current best-performing method, for both size-performance and speed-performance trade-offs. We present the results in Figure \ref{fig:tradeoff}. The left figure measures the trend of \method's performance after each replacing operation in the size-first assembling strategy.  The right figure illustrates the performance of \method and the KD baseline with configurations that correspond to a speed-up ratio from approximately $1.25\times$ to $2.5\times$. We can see that \method retains the performance of the original model very well when only reducing a few layers or assembling for a relatively small speed-up ratio. More importantly, \method consistently outperforms the KD baseline, which is trained 6 times (once for each specific compression ratio), and the improvement is even larger in the extreme compression regime. This confirms the effectiveness of our approach for flexible seq2seq compression and its robustness to relatively large compression ratios.

\subsection{Analysis}

We then conduct a series of analyses to better understand the effectiveness of \method. All experiments are conducted with T5-base on the CNN/DailyMail dataset.

\begin{figure}
\centering
\begin{minipage}{.25\textwidth}
  \centering
  \includegraphics[width=\textwidth]{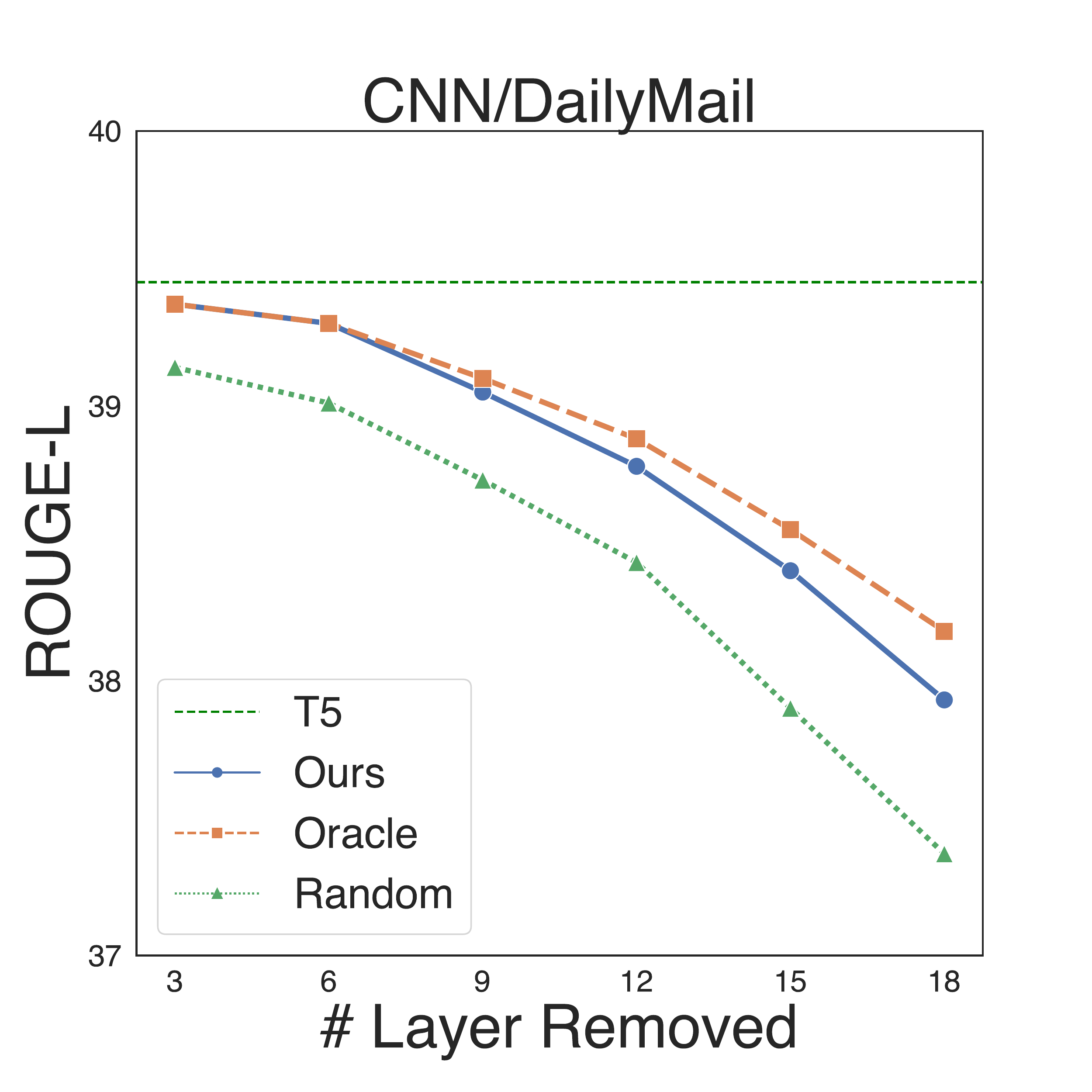}
  \label{fig:test1}
\end{minipage}%
\begin{minipage}{.25\textwidth}
  \centering
  \includegraphics[width=\textwidth]{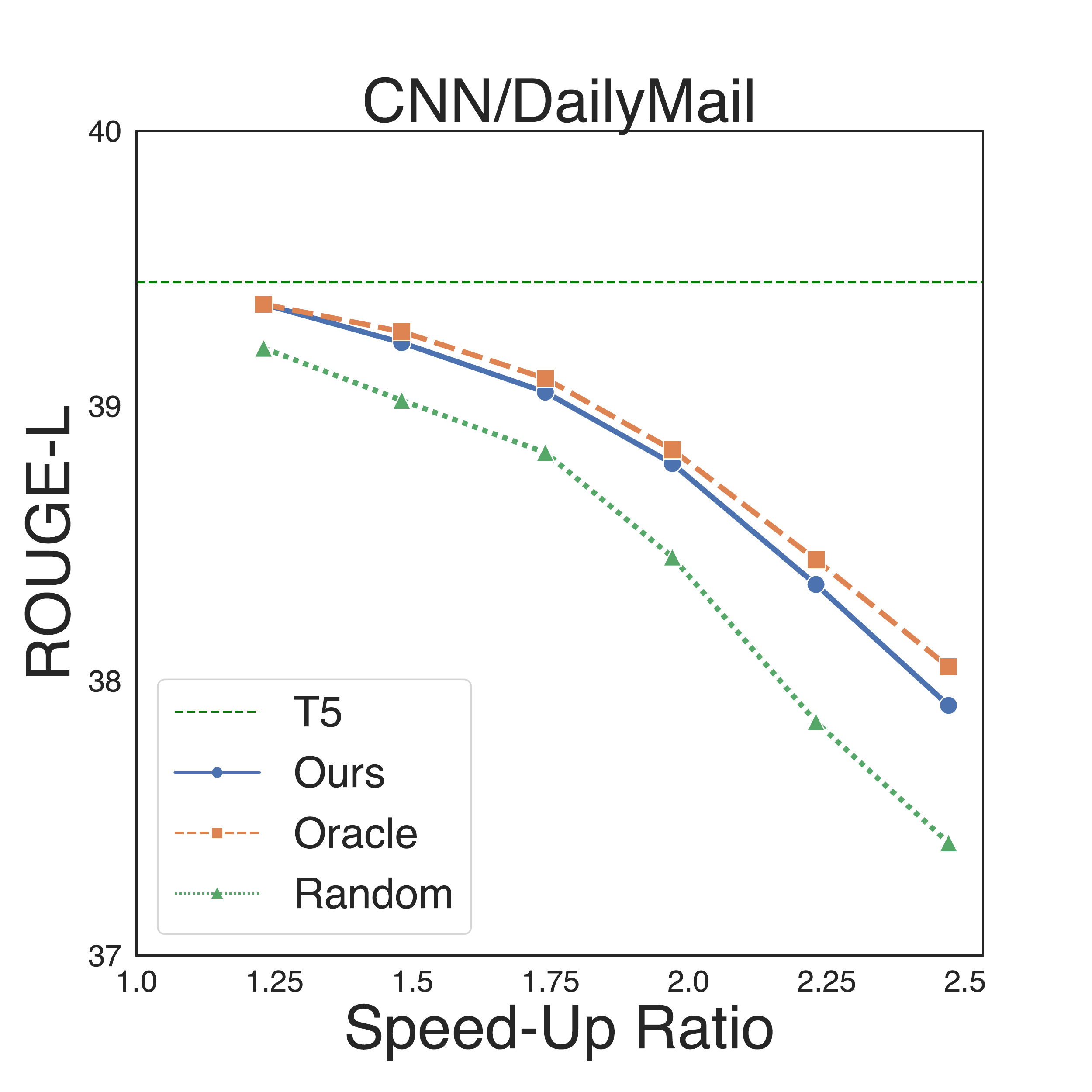}
  \label{fig:test2}
\end{minipage}
\vspace{-0.6cm}
\caption{Comparison between our flexible assembling method, random assembling baseline, and oracle performance with T5-base on the CNN/DailyMail dataset in both size-first (left) and speed-first (right) settings.}

\label{fig:strategy}
\end{figure}

\paragraph{Impact of Assembling Strategy}
We first analyze the impact of the proposed assembling strategies which replace layers from fine to coarse, from encoder to decoder, and from top to bottom. We compare our strategies in both the size-first and speed-first settings with the random baseline and the oracle where we exhaustively search all possibilities. The results are shown in Figure \ref{fig:strategy}. We can see that our strategy is relatively close to the oracle, especially in the low compression ratio regime. In contrast, the random assembling baseline performs substantially worse, demonstrating the effectiveness of our assembling strategies. 

\paragraph{Impact of Random Replacing}
Since the assembling strategy at test time is deterministic, one may question the need for a random module replacing scheme during training. To this end, we compare \method with the baseline where we only sample model structures used during inference. The results are shown in Table \ref{tab:analysis}. We can see that this variant performs substantially worse than the random replacing method. We believe that this is due to the interaction between well-trained original layers and the modularized layers and is crucial for improving performance. 

\paragraph{Impact of Knowledge Distillation}

We then study the impact of combining knowledge distillation with module replacing for training \method. We compare the variant trained without knowledge distillation in Table \ref{tab:analysis}. We find that incorporating knowledge distillation for training modularized layers improves the overall performance. This  confirms the complementary nature of knowledge distillation and module replacement for the first time.

\paragraph{Impact of Curriculum Replacing}
In addition, we also analyze the impact  of curriculum replacing which progressively increases the probability of including coarse-grain modularized layers in the hybrid model during training. We train a uniform sampling variant for comparison. In Table \ref{tab:analysis}, we find that the uniform variant performs worse than \method, demonstrating the effectiveness of curriculum replacing.

\begin{table}[]
    \centering
    \resizebox{\linewidth}{!}{
    \begin{tabular}{lccc}
    \toprule
         Methods & RG-1 & RG-2 & RG-L  \\
    \midrule
         Ours & \bf 41.53 & \bf 19.74 & \bf 38.62 \\
        - w/o random replacing & 41.25 & 19.52 & 38.31 \\
        - w/o curriculum replacing & 41.35 & 19.62 & 38.43 \\
        - w/o knowledge distillation & 41.42 & 19.60 & 38.49 \\
    \bottomrule
    \end{tabular}}
    \caption{Ablation study results with T5-base on the CNN/DailyMail dataset in the size-first setting. }
    \label{tab:analysis}
\end{table}

\section{Conclusion}


We introduce \method, a framework for flexible and accurate seq2seq compression that adapts to different performance-efficiency trade-offs, 
which has important practical implications such as reducing the computation cost and environmental impact of deployed systems. 
Our method defines a set of modularized layers with different granularity and trains them to share the same functionality as the sub-networks of the original model they replace. We combine multi-grained module replacing and knowledge distillation, and design two flexible assembling strategies for size-first and speed-first inference. Empirical results on various text generation tasks show that our approach consistently outperforms previous methods while alleviating the need to re-train the compact model in order to adapt to new memory and inference time constraints.

\section*{Limitations}

Our experiments focus on the T5-base and T5-large models as these are widely used, representative pre-trained seq2seq models. However, there are other pre-trained seq2seq models such as BART that we did not experiment with. It would also be interesting to experiment with pre-trained models with more layers such as T5-3B and T5-11B. We have not conducted these experiments due to resource constraints.

\section*{Ethics Statement}

Our method is used to compress seq2seq Transformer models. Therefore, ethical considerations of text generation models generally apply to our method. We encourage users to assess potential biases before deploying text generation models.

\bibliography{anthology,custom}
\bibliographystyle{acl_natbib}

\appendix



\end{document}